\documentclass[letterpaper, 10 pt, conference]{ieeeconf}
\IEEEoverridecommandlockouts
\usepackage{graphicx}
\def\BibTeX{{\rm B\kern-.05em{\sc i\kern-.025em b}\kern-.08em
    T\kern-.1667em\lower.7ex\hbox{E}\kern-.125emX}}
\usepackage[font=small,labelfont=bf,tableposition=top]{caption}
\usepackage{blindtext}
\usepackage{xspace}
\usepackage{xcolor}
\usepackage{graphicx}     
\usepackage{pifont}
\usepackage{amsmath}
\usepackage{amssymb}
\usepackage{booktabs}
\usepackage{multirow}
\usepackage{paralist}
\usepackage{fontawesome}
\usepackage{algorithm}
\usepackage{algorithmic}

\definecolor{dcolor}{RGB}{204,229,255}
\definecolor{pinkregion}{RGB}{255,182,193}
\definecolor{BlueRegion}{RGB}{0,0,205}  
\definecolor{OrangeRegion}{RGB}{230,120,50}
\definecolor{YellowRegion}{RGB}{182,140,2} 
\definecolor{LSDMon}{RGB}{172,215,142} 
\definecolor{Visual}{RGB}{254,217,97}
\usepackage[table]{xcolor}

\definecolor{hollywoodcerise}{rgb}{0.96, 0.0, 0.63}
\definecolor{lasallegreen}{rgb}{0.03, 0.47, 0.19}
\definecolor{hanpurple}{rgb}{0.32, 0.09, 0.98}
\definecolor{green(pigment)}{rgb}{0.0, 0.65, 0.31}

\newcommand{\ie}{\emph{i.e.}\xspace}

\makeatletter
\let\NAT@parse\undefined
\makeatother

\usepackage[pagebackref=false,breaklinks=true,colorlinks,bookmarks=false]{hyperref}
\hypersetup{colorlinks=true,linkcolor={red},citecolor={hanpurple},urlcolor={magenta}}

\begin{document}

\pagestyle{empty}

\title{\LARGE \bf
Hallucinating 360{\textdegree}: Panoramic Street-View Generation via Local Scenes Diffusion and Probabilistic Prompting
}

\author{Fei Teng$^{1,*}$, Kai Luo$^{1,*}$, Sheng Wu$^{1,*}$, Siyu Li$^{1}$, Pujun Guo$^{1}$, Jiale Wei$^{2}$,\\Jiaming Zhang$^{1}$, Kunyu Peng$^{2,3}$, and Kailun Yang$^{1,\dag}$
\thanks{This work was supported in part by the National Natural Science Foundation of China (Grant No. 62473139), in part by the Hunan Provincial Research and Development Project (Grant No. 2025QK3019), and in part by the State Key Laboratory of Autonomous Intelligent Unmanned Systems (the opening project number ZZKF2025-2-10).}
\thanks{$^{1}$F. Teng, K. Luo, S. Wu, S. Li, P. Guo, J. Zhang, and K. Yang are with the School of Artificial Intelligence and Robotics and the National Engineering Research Center of Robot Visual Perception and Control Technology, Hunan University, Changsha 410082, China (email: kailun.yang@hnu.edu.cn).}
\thanks{$^{2}$J. Wei and K. Peng are with the Institute for Anthropomatics and Robotics, Karlsruhe Institute of Technology, 76131 Karlsruhe, Germany.}
\thanks{$^{3}$K. Peng is also with INSAIT, Sofia University ``St. Kliment Ohridski'', Sofia 1784, Bulgaria.}
\thanks{$^{\dag}$Corresponding author: Kailun Yang.}
}

\maketitle
\thispagestyle{empty}

\begin{abstract}
Panoramic perception holds significant potential for autonomous driving, enabling vehicles to acquire a comprehensive 360{\textdegree} surround view in a single shot. However, autonomous driving is a data-driven task. Complete panoramic data acquisition requires complex sampling systems and annotation pipelines, which are time-consuming and labor-intensive. Although existing street view generation models have demonstrated strong data regeneration capabilities, they can only learn from the fixed data distribution of existing datasets and cannot leverage stitched pinhole images as a supervisory signal. In this paper, we propose the first panoramic generation method \textbf{Percep360} for autonomous driving. Percep360 enables coherent generation of panoramic data with control signals based on the stitched panoramic data. Percep360 focuses on two key aspects: coherence and controllability. Specifically, to overcome the inherent information loss caused by the pinhole sampling process, we propose the Local Scenes Diffusion Method (LSDM). LSDM reformulates the panorama generation as a spatially continuous diffusion process, bridging the gaps between different data distributions. Additionally, to achieve the controllable generation of panoramic images, we propose a Probabilistic Prompting Method (PPM). PPM dynamically selects the most relevant control cues, enabling controllable panoramic image generation. We evaluate the effectiveness of the generated images from three perspectives: image quality assessment (\ie, no-reference and with reference), controllability, and their utility in real-world Bird's Eye View (BEV) segmentation. Notably, the generated data consistently outperforms the original stitched images in no-reference quality metrics and enhances downstream perception models, leading to an improvement of $2.5\%$ in mIoU for panoramic BEV segmentation. The source code will be publicly available at \url{https://github.com/FeiT-FeiTeng/Percep360}.
\end{abstract}

\section{Introduction}

Autonomous vehicles require the integration of surrounding environmental information to achieve 360-degree perception. In this context, panoramic cameras have garnered increasing attention from the research community due to their large Field of View (FoV)~\cite{AutoPano,PanoramicReview2,ai2025survey}. Numerous studies have employed panoramic cameras for downstream perception tasks for street views, such as optical flow estimation~\cite{Panoflow,Panoflow2}, object detection~\cite{PanoOD,Ominitrack}, and semantic segmentation~\cite{AutoPano,PanoSS,PanoSS3}. 
However, as a typical data-driven task, developing safe and effective autonomous driving systems heavily relies on the scale and diversity of the collected datasets. Due to the challenges in constructing panoramic data acquisition and annotation systems, the development of panoramic autonomous driving has been hindered by the scarcity of large-scale datasets. 
OneBEV~\cite{Onebev} has attempted to use stitching-based methods to project pinhole images onto a spherical surface, thus simulating panoramic distortion, avoiding the need for data acquisition and offering a new feasible paradigm for panoramic autonomous driving. However, the stitching methods are unable to produce an infinite number of images from prompt input and fail to address the information misalignment caused by the sampling process.

\begin{figure}[!t]
    \centering
    \includegraphics[width=0.48\textwidth]{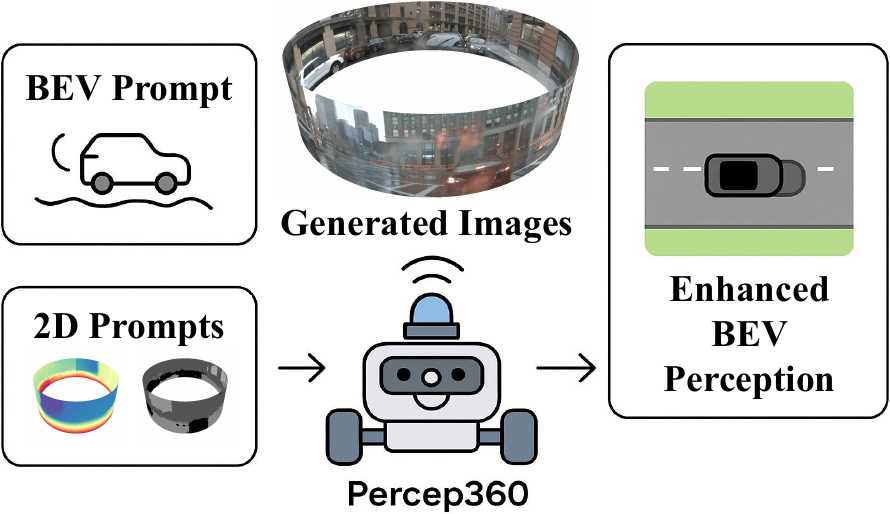}
    \captionof{figure}{Percep360 leverages diverse prompts to generate coherent panoramic images. 
    The generated images exhibit improved quality and controllability. As a data generator, Percep360 synthesizes diverse data to avoid the expensive process of data processing, thereby enhancing the performance of semantic mapping.
    }

    \vspace{-2em}
\end{figure}

\begin{figure}[!t]
    \centering
    \includegraphics[width=0.48\textwidth]{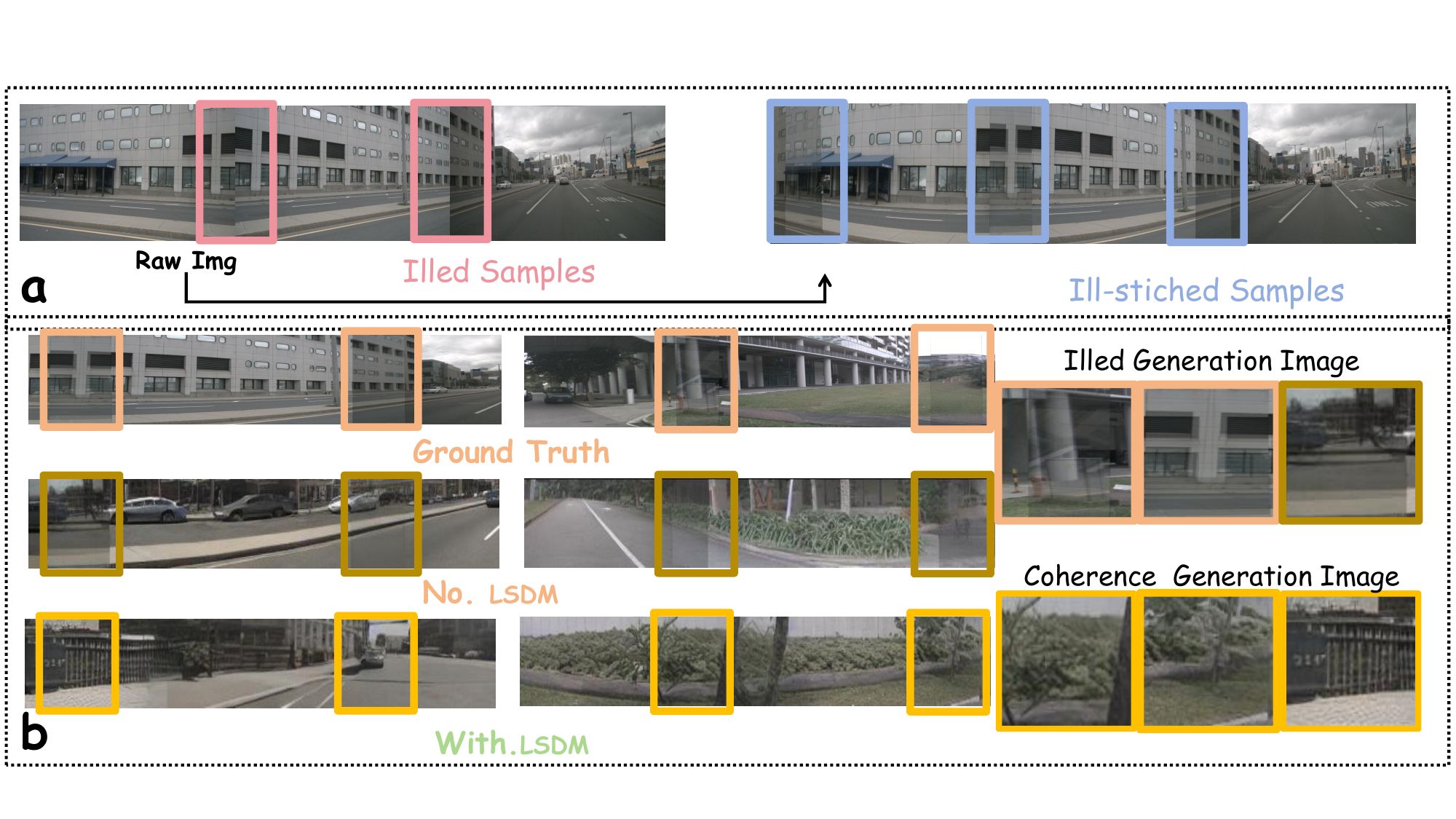}
    \caption{Visualizations of the \textcolor{pinkregion}{pinhole samples}, \textcolor{BlueRegion}{stitched image}, \textcolor{OrangeRegion}{ground truth} for training generation model, and \textcolor{YellowRegion}{the results of the existing approach} are presented. Owing to the inherent information loss and misalignment in pinhole camera sampling, \textcolor{BlueRegion}{stitching-based methods} struggle to produce coherent panoramic images. Moreover, when these \textcolor{OrangeRegion}{stitched images} are used as inputs, \textcolor{YellowRegion}{existing generation models} often inherit the stitching errors. In contrast, \textcolor{LSDMon}{the proposed LSDM approach} enables the generation of images with improved coherence, particularly at stitching boundaries.}
    \vspace{-1.7em}
    \label{fig:Fig2}
\end{figure}

The recent success of generative models for pinhole cameras has introduced an economic paradigm to avoid the high costs associated with data collection and annotation. Several models~\cite{ControlNet,Controlnext,BevGen,Bevcontrol} have proposed generation frameworks trained on prompt-image pairs, enabling fine-grained control over the diffusion process. In a similar vein, several works~\cite{Drive123,Occword,EdiDriving,MD} have emerged, offering cost-effective solutions for closed-loop autonomous driving data generation. 
However, existing methods that utilize pinhole samples to facilitate scene and style transfer in driving scenarios fail to ensure coherence in panoramic image synthesis. 
We remodify the existing models for the panoramic domain. Those networks learn the distribution of stitched images, including misalignment caused by the stitching process and inherent information loss introduced by the pinhole sampling process, as shown in Fig.~\ref{fig:Fig2}. 
This can lead to perception errors and localization inaccuracies, further impairing the accuracy of environmental modeling and ultimately affecting the practical applications of robots.

\begin{center}
\colorbox{gray!15}{
  \parbox{0.9\linewidth}{
    \centering
    \faCar~\textbf{\textit{Thus, effectively generating coherent panoramic images remains a crucial challenge for panoramic autonomous driving systems.}}
  }
}
\end{center}

To address this problem, we propose the first solution: \textbf{Percep360}. Percep360 achieves coherent panoramic data generation by leveraging stitched pinhole images as a supervisory signal. As shown in Fig.~\ref{fig:Fig2}(a), due to sampling errors from cameras, spatial misalignments occur when stitching multiple images into a panoramic view. Furthermore, we use BEVControl~\cite{Bevcontrol} as the baseline for controllable image generation. The generated images inherit the sampling errors and fail to produce consistent panoramic views. Meanwhile, since there is no supervisory guidance for coherence in the stitched region, the network inevitably alters the data distribution in that area when attempting to achieve coherence, which inevitably reduces the controllability of the image. 
From this perspective, \ding{192} to achieve coherence in panoramic images, we employ \textit{Local Scenes Diffusion Method} (LSDM). LSDM circularly adjusts the correspondence between the network and the image-prompt pair data, leveraging locally coherent image features to bridge the information loss inherent to the pinhole sample. Furthermore, unlike panoramic image generation, street view synthesis must preserve controllability while maintaining image quality. As shown in Fig.~\ref{fig:Fig2}(b), the spatial positions of roads and flower beds have changed. To address the decline in controllability resulting from compensation for sample errors, \ding{193} we introduce \textit{Probabilistic Prompting Method} (PPM), which allows dynamic adjustment of prompt feature cues across different spatial regions, thereby achieving controllable generation. We evaluate the generated images on the nuScenes-360 dataset~\cite{Onebev} from three perspectives: image quality assessment, no-reference image quality assessment, and controllability. Our method is the first to specifically target panoramic street-view image generation. It consistently improves both image quality and controllability, while boosting panoramic BEV segmentation performance by $2.5\%$ in mIoU.

At a glance, we deliver the following contributions:
\begin{compactitem}
    \item This paper proposes Percep360, the first panoramic generation framework for autonomous driving. Percep360 enables the generation of coherent and controllable panoramic data using inconsistent stitched imagery.
    \item To fulfill the inherent information loss caused by the pinhole sampling process, we introduce the Local Scenes Diffusion Method (LSDM). LSDM leverages local information to realize a spatially continuous diffusion process, generating coherent panoramic scenes.
    \item To mitigate the controllability drop that arises when compensating for sample errors, we propose a Probabilistic-Prompt Method (PPM). PPM adjusts the relationships among visual, textual, and other prompts, resulting in controllable panoramic generation.
    \item Our experiments demonstrate that Percep360 enables coherent and controllable panoramic scene generation. Moreover, the synthetic data it produces leads to an improvement of $2.5\%$ in the performance of the downstream perception task of panoramic BEV segmentation.
\end{compactitem}

\section{Related Work}
To address the scarcity of data in panoramic autonomous driving, we explore a cost-effective data synthesis paradigm from the perspective of driving scenes generation. Furthermore, to achieve coherent data generation and compensate for stitched image errors, we investigate existing methods for panoramic image synthesis.

\textbf{Driving Scenes Generation.} 
Autonomous driving is a data-driven task, and data collection is labor-intensive. Leveraging generative paradigms to address information scarcity in the autonomous driving process has become an increasingly popular approach in recent research. BEVGen~\cite{BevGen} utilizes BEV maps to generate multi-view images, while BEVControl~\cite{Bevcontrol} introduces a two-stage training paradigm to enhance the distinction between foreground and background, thereby improving the accuracy of generated images. Based on that, models such as DriveScape~\cite{DriveScape}, DrivingDiffusion~\cite{DrivingDiffusion}, HoloDrive~\cite{holodrive}, and MagicDrive~\cite{MD} introduce bounding boxes to further enhance the controllability of the generated content. Meanwhile, Text2Street~\cite{Tex2St} leverages object and road topology prompts to achieve controllable single-image generation. 
Additionally, several works~\cite{panacea,Magicdrivedit,Perldiff,UniMLVG,Stag1} propose the generation of multi-view videos using cross-frame methods. \textit{However}, these works focus on generating controllable multi-view pinhole images under data-rich and clean conditions. In contrast, our goal is to generate coherent and controllable panoramic street-view images using imperfectly stitched images.

\begin{figure*}[ht]
    \centering
    \includegraphics[width=0.98\textwidth]{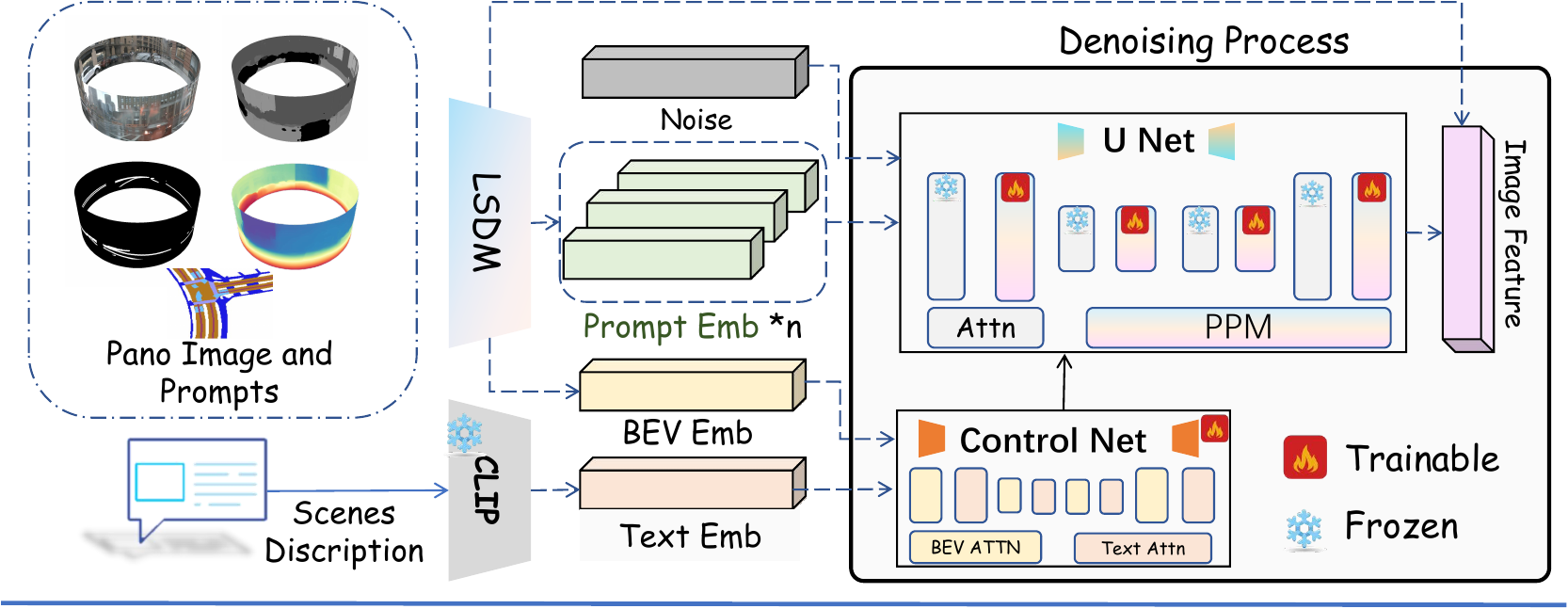} 
    \caption{The overall architecture of Percep360 leverages the BEV map, textual prompt, depth map, and mask map jointly as guidance signals. With the integration of the LSDM and PPM modules, the framework achieves coherent and controllable panoramic generation.}
    \label{fig:architecture}
    \vspace{-1.5em}
\end{figure*}

\textbf{Panoramic Image Synthesis.} 
The synthesis of panoramic images can be categorized into two aspects: text-based generation and completion based on partial image information. \textbf{Generation:} Text-to-image synthesis~\cite{Dreammatcher,PALP,OPT2I,Raphael} has rapidly emerged as a prominent area of research. For instance, PanFusion~\cite{Panfusion} employs a dual-branch architecture to seamlessly integrate additional constraints, enabling more customizable and high-quality panoramic outputs. Several works~\cite{PanoFree,PanoGen,360dvd} achieve panoramic image completion by extending the field of view, enabling seamless and realistic scene expansion. \textbf{Completion:} The image inpainting task aims to restore missing regions of an image while maintaining overall coherence. PICFormer~\cite{PICFormer} introduces the Codebook-based sharing paradigm and enhances restoration performance through plus-realism image completion~\cite{PIC}. RePaint~\cite{RePaint} models the semantics and texture of missing regions by leveraging a denoising diffusion probabilistic model. BrushNet~\cite{Brushnet,Brushedit} incorporates a mask branch to better distinguish masked regions. Several works~\cite{Panodiff1,PanoDiff2} introduce the rotation-based paradigm to address the continuity challenges in panoramic image completion, ensuring smoother and more coherent results.
\textit{However}, a fundamental limitation exists: These works are trained using consistent images. Such a paradigm is not suitable for generating coherent panoramic data from imperfect samples. Moreover, existing panoramic image synthesis methods utilize weak textual prompts, which fail to meet the requirements of autonomous driving for accurate road topology representation.

\section{Methodology}
In this section, we first introduce the basic process of diffusion and present the problem formulation of our work in Sec.~\ref{Sec.IIIA}. Then, in Sec.~\ref{Sec.IV}, we focus on describing the overall architecture of our network and the proposed method. 

\subsection{Preliminary and Problem Formulation}\label{Sec.IIIA}
\noindent\textit{Preliminary:} Latent Diffusion Models (LDM)~\cite{LDM} represent traditional diffusion models by conducting the diffusion process within a compact latent representation space, typically involving a two-step procedure: feature compression and iterative denoising for modeling data distributions. In the feature compression, input images in the pixel space are first transformed into compact latent embeddings using a perceptual compression model $E$, significantly reducing the complexity associated with processing high-dimensional data: $z = E(x)$. This latent representation $z$  passes through the diffusion process and can be decoded back into the original image domain using the decoder $D$, formally expressed as: $\hat{x} = {D}(z)$. 
In the iterative denoising for modeling data distributions, the diffusion process is carried out within this latent space. In the forward diffusion stage, gaussian noise is gradually added to the latent embeddings, progressively transforming the latent representation into a distribution close to a standard Gaussian: $z_t = \sqrt{\bar{\alpha}_t} z + \sqrt{1 -\bar{\alpha}_t}\epsilon,\enspace \epsilon \sim \mathcal{N}(0, I)$. Here, $\bar{\alpha}_t$ is a pre-defined noise scheduling parameter, and $t$ denotes the diffusion timestep. 
In the reverse denoising stage, a U-Net is trained to estimate and remove the added noise from the noisy latent representation $z_t$. This network is parameterized as $\epsilon_\theta$, trained by minimizing the following mean-square error loss:
$\min_{\theta} \mathbb{E}_{z, c, \epsilon, t}\left[\|\epsilon - \epsilon_{\theta}(z_t, t, c)\|^2\right]$, where $c$ represents conditional information, guiding the conditional image generation. After training, the LDM can generate samples by iterative denoising a latent feature initialized from Gaussian noise. Formally, this iterative sampling process can be expressed as: $p_{\theta}(z_{t-1}|z_t, c) = \mathcal{N}(z_{t-1}; \mu_{\theta}(z_t, t, c), \Sigma_{\theta}(z_t, t, c))$, where $\mu_{\theta}$ and $\Sigma_{\theta}$ are predicted by the denoising network. 

\begin{figure*}[!t]
    \centering
    \includegraphics[width=0.98\textwidth]{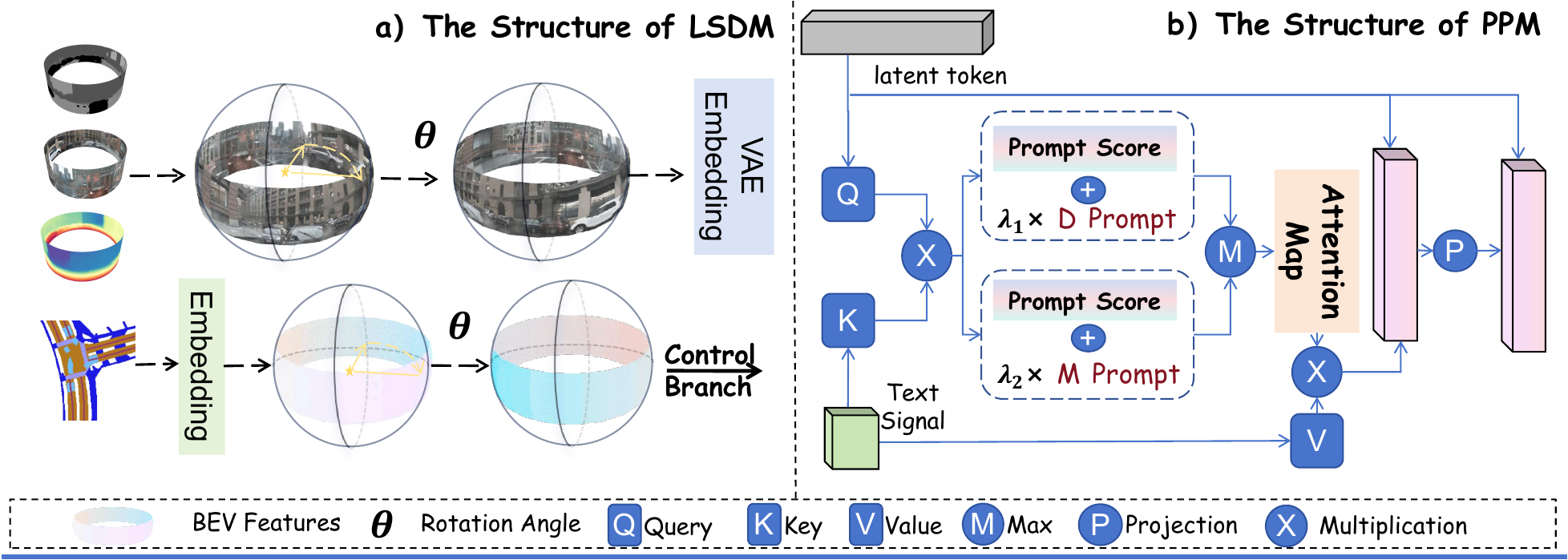} 
    \caption{The architecture of LSDM and PPM is illustrated in Figures (a) and (b), respectively. By applying circular rotations to diverse prompts and BEV features, LSDM reformulates panoramic image generation as a spatially continuous diffusion process. PPM further enhances controllability by dynamically selecting the most relevant control cues, leading to an accurate and consistent panoramic image.}
    \label{fig:panorama}
    \vspace{-1.5em}
\end{figure*}

\vspace{0.3em}
\noindent\textit{Problem Formulation:} 
Given a panoramic image defined over a pixel domain \(\Omega \subset \mathbb{R}^2\), the entire data distribution of the image is denoted by \(f(x)\), where \(x \in \Omega\). The panoramic image is constructed by stitching multiple pinhole camera views, resulting in a composite pixel space partitioned into twelve regions: six coherent regions \(\{\mathcal{R}_i\}_{i=1}^6\) and six aliased regions \(\{\mathcal{A}_i\}_{i=1}^6\). Thus, the domain can be expressed as: $\Omega = \left( \bigcup_{i=1}^6 \mathcal{R}_i \right) \cup \left( \bigcup_{i=1}^6 \mathcal{A}_i \right), \enspace \mathcal{R}_i \cap \mathcal{A}_j = \emptyset, \enspace \forall i, j$. In the initial state, the distribution of the image exhibits regional coherence. The local distribution over each coherent region \(\mathcal{R}_i\), denoted \(f_i(x)\), possesses strong spatial consistency. In contrast, the distribution over each aliased region \(\mathcal{A}_i\), denoted \(\tilde{f}_i(x)\), is complex and discontinuous due to stitching artifacts. Our objective is to enhance the model's ability to capture the structure of the coherent regions by introducing an operation \(R_{\theta}(\cdot)\). Meanwhile, by incorporating a control method, the network leverages prompt signals \(A\) and \(B\) to generate a controllable distribution: $
f(x \mid A, B) \approx \sum_{i=1}^6 \mathbb{I}_{\mathcal{R}_i}(x) \cdot f_i(x \mid A, B)$, where \(\mathbb{I}_{\mathcal{R}_i}(x)\) is the indicator function for region \(\mathcal{R}_i\). Specifically, we seek to transform the initial discontinuous mixed distribution: $f(x) = \sum_{i=1}^6 \mathbb{I}_{\mathcal{R}_i}(x) \cdot f_i(x) + \sum_{i=1}^6 \mathbb{I}_{\mathcal{A}_i}(x) \cdot \tilde{f}_i(x)$, into a coherent, structure-dominant distribution: $f(x \mid A, B) \longrightarrow \tilde{f}_{\text{coherent}}(x), \enspace x \in \Omega$.

\subsection{The Structure of Percep360}\label{Sec.IV}
In this paper, we propose Percep360, the first paradigm for achieving high-quality controllable panoramic image generation. Specifically, Percep360 adopts a side-controlling architecture, following~\cite{ControlNet,Controlnext}. The network introduces LSDM, enabling more coherent image generation under the panorama stitching setting. Furthermore, we introduce the PPM module, which injects prompt embeddings into the global features of panoramic images to enable controllable image generation.

\noindent\textit{Overall Architecture:} 
As shown in Fig.~\ref{fig:architecture}, the control branch of U-Net processes the guidance signals, and the resulting features are injected into the main U-Net backbone to condition the generation. Specifically, the model takes as input three 2D data: the RGB image $\mathbf{I}$, the binary mask $\mathbf{M}$, and the depth map $\mathbf{D}$.
In addition, the input includes a BEV map $\mathbf{B} \in \mathbb{R}^{B \times K \times 200 \times 200}$ and a text prompt $\mathbf{T}$. Each of the 2D inputs---RGB image, depth map, and mask---is individually encoded using a frozen variational autoencoder encoder $\mathcal{E}_{\text{VAE}}$ with shared weights. This yields the following latent features:
\begin{equation}
\mathbf{F}_{\text{img,depth,mask}} = \mathcal{E}_{\text{VAE}}(\mathbf{I}, \mathbf{D}, \mathbf{M}).
\end{equation}
Next, the BEV map is encoded using a dedicated BEV encoder $\mathcal{E}_{\text{BEV}}$, following BEVControl~\cite{Bevcontrol}. The resulting feature map is resized to match the spatial resolution of the 2D features, as shown in the following equation:
\begin{equation}
\mathbf{F}_{\text{BEV}} = \text{Interp}\left( \mathcal{E}_{\text{BEV}}\left( \mathbf{B} \right),\ \text{size}=(h, w) \right).
\end{equation}
Meanwhile, the text prompt $\mathbf{T}$ is processed through a CLI text encoder $\mathcal{E}_{\text{CLIP}}$~\cite{clip}, producing a semantic feature vector:
\begin{equation}
\mathbf{F}_{\text{TXT}} = \mathcal{E}_{\text{CLIP}}(\mathbf{T}) \in \mathbb{R}^{B \times d}.
\end{equation}
Finally, the three encoded components (\ie, $\mathbf{F}_{2D}$, $\mathbf{F}_{\text{BEV}}$, and $\mathbf{F}_{\text{TXT}}$) are fed into the control branch. This branch modulates the intermediate feature maps at multiple scales, enabling spatially and semantically controllable feature generation.

\noindent\textit{The design of LSDM:} 
To achieve more coherent panoramic generation, we propose the Local Scenes Diffusion Method (\textbf{LSDM}), as shown in Fig.~\ref{fig:panorama}, which mitigates the discontinuities at image seams. Since a panoramic image spans a full $360^\circ$ horizontal field of view, we treat its domain as a spherical manifold where the left and right boundaries are adjacent. For the 2D inputs, including the image $\mathbf{I}$, mask $\mathbf{M}$, and depth map $\mathbf{D}$, we perform a horizontal wrap-around rotation before feeding them into the network. A rotation angle $\theta \in [0, 2\pi)$ is randomly sampled and converted into a pixel displacement:
\begin{equation}
\Delta w = \left\lfloor \frac{\theta}{2\pi} \cdot W \right\rfloor,
\end{equation}
where $W$ is the width of the panoramic image. The rotated result is obtained via a circular shift:
\begin{equation}
\mathbf{X}_{\text{rot}} = \text{Roll}(\mathbf{X},\ \Delta w), \quad \mathbf{X} \in \{\mathbf{I}, \mathbf{M}, \mathbf{D}\},
\end{equation}
where the Roll operator is defined as:
\begin{equation}
\text{Roll}(\mathbf{X},\ \Delta w)[u, v] = \mathbf{X}\left[(u + \Delta w) \bmod W,\ v\right],
\end{equation}
with $\mathbf{X}$ denoting the input 2D tensor (image, mask, or depth map), $W$ the horizontal width of $\mathbf{X}$, $(u,v)$ the horizontal and vertical pixel coordinates, and $\bmod$ the modulo operator ensuring wrap-around indexing.

The BEV input $\mathbf{B}$ is encoded into a feature map:
\begin{equation}
\mathbf{F}_{\text{BEV}} = \mathcal{E}_{\text{BEV}}(\mathbf{B}).
\end{equation}

Then we apply a similar circular shift in the feature space using the same angle $\theta$:
\begin{equation}
\Delta w_{\text{BEV}} = \left\lfloor \frac{\theta}{2\pi} \cdot W_{\text{BEV}} \right\rfloor,
\quad
\mathbf{F}_{\text{BEV}}^{\text{rot}} = \text{Roll}(\mathbf{F}_{\text{BEV}},\ \Delta w_{\text{BEV}}).
\end{equation}

This feature-space rotation is consistent with the 2D image rotation, but applied directly to the encoded BEV representation. By synchronizing the angular shift across both 2D and BEV modalities, LSDM facilitates geometrically aligned and spatially consistent panoramic generation.

\begin{table*}[htbp]
\setlength{\abovecaptionskip}{3pt} 
\setlength{\belowcaptionskip}{0pt}
\setlength{\tabcolsep}{2mm}
\centering
\small
\vspace{0.5em}

\renewcommand{\arraystretch}{1.1}
\resizebox{0.98\textwidth}{!}{
    \begin{tabular}{@{}l|cc|cc|cc|c@{}}
        \toprule
        & \multicolumn{2}{c|}{NRIQA} & \multicolumn{2}{c|}{IQA} & \multicolumn{2}{c|}{Control} & Rank.\\     
        \cmidrule(lr){2-3}\cmidrule(lr){4-5}\cmidrule(lr){6-7}
        Method & BRISQUE~\cite{BRISQUE} ($\downarrow$)  & PIQE ($\downarrow$) & SSIM~\cite{SSIM} ($\uparrow$) &  FID~\cite{FID} ($\downarrow$) & Driv. ($\uparrow$) & Mean. ($\uparrow$) & ($\downarrow$) \\
        \midrule
        Oracle   & 19.39 & 12.06     & \cellcolor{gray!30} & \cellcolor{gray!30}  & 0.59& 0.47& N/A \\     
        MagicDrive*~\cite{MD} & 20.65 (2) & 12.13 (2) & 0.16 (1) & 14.16 (1) & 0.22 (2) & 0.12 (2) & 10 (2)\\

        Baseline~\cite{Bevcontrol} & 22.40 (3) & 14.32 (3) & 0.14 (3) & 22.07 (3) & 0.21 (3) & 0.12 (2) & 17 (3)\\

        Percep360 & 20.24 (1) & 11.44 (1) & 0.16 (1) & 14.43 (2) & 0.25 (1) & 0.13 (1) & 7 (1)\\
        \midrule \midrule        
    \end{tabular}
}
\caption{Comparison with state-of-the-art methods on the nuScenes validation set in \textit{Generation Quality}. $\uparrow / \downarrow$ indicates that a higher/lower value is better. `Oracle' presents the row results of corresponding models trained on the nuScenes training set. $^*$ indicates that the model has undergone structural modifications to accommodate the panoramic image generation task.}
\label{tab:tab1}
\vspace{-1em}
\end{table*}

\noindent\textit{The design of PPM:} 
To enhance scene-level controllability and geometric consistency in diffusion-based generation, we design the Probabilistic Prompting Method (PPM) module that incorporates structural features into image features. This module dynamically modulates attention based on two forms of masking: depth map and mask map for the entire image. Given the depth map \(\mathbf{M}_d \in \mathbb{R}^{H_d \times W_d}\) and the mask map \(\mathbf{M}_m \in \mathbb{R}^{H_m \times W_m}\), we first interpolate and reshape them to obtain spatially aligned maps \(\tilde{\mathbf{M}}_d, \tilde{\mathbf{M}}_m \in \mathbb{R}^{H \times W}\), matching the spatial resolution of the visual token sequence \(x \in \mathbb{R}^{N \times d}\), where \(N = H \times W\):
\begin{equation}
\tilde{\mathbf{M}}_d = \text{Interpolate}(\mathbf{M}_d), \quad \tilde{\mathbf{M}}_m = \text{Interpolate}(\mathbf{M}_m).
\end{equation}

The text signal \(c \in \mathbb{R}^{L \times d}\) is linearly projected to the key and value spaces:
\begin{equation}
k = W_k c \in \mathbb{R}^{L \times d}, \quad v = W_v c \in \mathbb{R}^{L \times d},
\end{equation}
where \(W_k, W_v \in \mathbb{R}^{d \times d}\) are learnable projection matrices. The latent token sequence \(x\) serves as the query after projection:
\begin{equation}
q = W_q x \in \mathbb{R}^{N \times d}.
\end{equation}

The scaled dot-product attention logits \(\mathbf{A} \in \mathbb{R}^{N \times L}\) are computed as:
\begin{equation}
\mathbf{A} = \frac{q k^\top}{\sqrt{d}}.
\end{equation}

We then inject spatial priors from depth and mask prompts by reshaping \(\tilde{\mathbf{M}}_d, \tilde{\mathbf{M}}_m\) into compatible matrices \(\mathbf{M}_d', \mathbf{M}_m' \in \mathbb{R}^{N \times L}\) and adding them with learnable scaling factors \(\lambda_1, \lambda_2 \in \mathbb{R}\):
\begin{equation}
\mathbf{A}_d = \mathbf{A} + \lambda_1 \mathbf{M}_d', \quad \mathbf{A}_m = \mathbf{A} + \lambda_2 \mathbf{M}_m'.
\end{equation}

The final attention map is obtained by element-wise maximum:
\begin{equation}
\mathbf{A}_{\text{final}} = \max(\mathbf{A}_d, \mathbf{A}_m).
\end{equation}

After normalization with softmax along the key dimension:
\begin{equation}
\text{Attn} = \text{softmax}(\mathbf{A}_{\text{final}}),
\end{equation}
the output representation is computed via a weighted sum over the values, followed by a linear projection and residual connection, following~\cite{LDM}:
\begin{equation}
x_{\text{out}} = \text{Proj}(\text{Attn} \cdot v) + x.
\end{equation}

\begin{figure*}[htbp]
    \centering
    \includegraphics[width=0.98\textwidth]{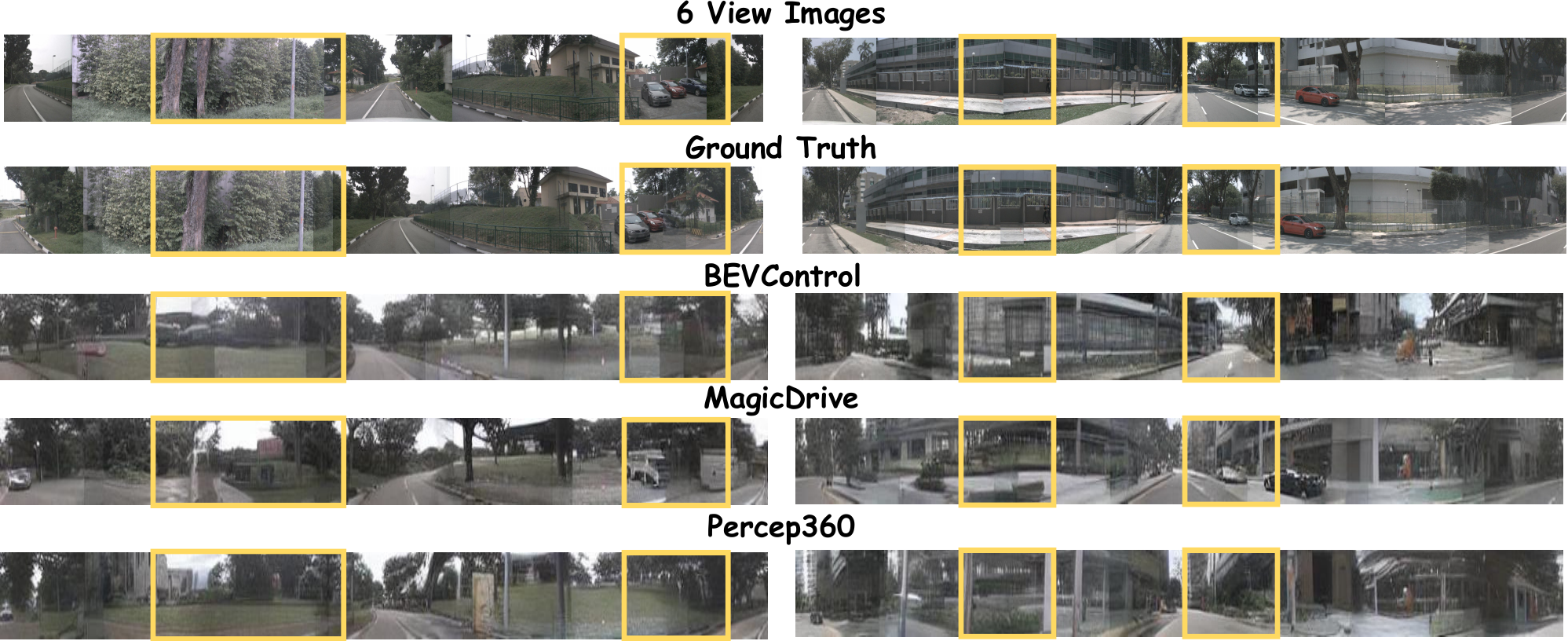}
    \caption{Visualizations of the generation results for the ground truth, BEVControl~\cite{Bevcontrol}, and MagicDrive~\cite{MD} are provided. Regions of stitching misalignment are highlighted with \textcolor{Visual}{yellow boxes}. It can be observed that, due to the lack of coherent generation capability, existing methods produce discontinuous results. In contrast, our method achieves improved coherence without compromising controllability.}
    \label{fig:Fig4}
    \vspace{-1.5em}
\end{figure*}

This attention mechanism enhances the controllability of the network and mitigates the controllability degradation caused by the continuous diffusion process.

Benefiting from the proposed design, our method enhances the quality of generated images and has the potential to boost the performance of downstream panoramic BEV perception.

\section{Experiments}

In this section, we present the experimental configuration in Sec.~\ref{Sec.IVA}, and then detail the quantitative and visualization results in Sec.~\ref{Sec.IVB}.
\subsection{Experiments Details}\label{Sec.IVA}
\noindent\textit{Dataset and Baselines.} The experiments are conducted on the nuScenes-360 dataset~\cite{Onebev}, which consists of panoramic driving scenes stitched from the nuScenes dataset~\cite{Nuscenes}. Following the official configuration, the dataset comprises $28,130$ images for training and $6019$ images for validation. We adopt BEVControl~\cite{Bevcontrol} as the baseline method. In our experiments, we replace the center view sample with a panoramic image without utilizing images from additional views.

\noindent\textit{Evaluation Metrics and Strategies.} We employ multiple evaluation strategies to assess the effectiveness of our panorama generation method comprehensively. Specifically, we utilize Fr\'{e}chet Inception Distance (FID) \cite{FID}, Structural Similarity Index (SSIM)~\cite{SSIM} for image quality assessment. Meanwhile, we also employ no-reference image quality assessment (\ie, BRISQUE~\cite{BRISQUE}, PIQE~\cite{PIQE}) to evaluate the quality by comparing the generated images with ground-truth images and those from other methods. Additionally, we evaluated the images, assessing their IoU score for road segmentation and mean intersection over union on the pre-trained BEV semantic segmentation network. Furthermore, to more intuitively compare the performance of different methods, we ranked the results for each score and obtained a ranking score.

\noindent\textit{Experiment Setting.} As we are the first to propose the generation of panoramic images using stitched pinhole images, addressing the current challenges in image collection and annotation within the panoramic generation community, we modify the BEVControl~\cite{Bevcontrol} to serve as our baseline. Both the image and control signal are represented in 2D format (\ie, BEV Map, Mask Map, Depth Map). Meanwhile, we introduce more comparisons with the 3D generation model, Magicdrive~\cite{MD}. Our experiments were conducted on two A6000 GPUs over a training period of ten days. We set the diffusion step to $20$ without using data augmentation strategies. To accelerate the diffusion process, the image is processed at a resolution of (60, 1200) following \cite{DriveSim}.

\vspace{-0.5em}
\subsection{Main Results}\label{Sec.IVB}
As shown in Tab.~\ref{tab:tab1}, we extend six-view image generation methods~\cite{Bevcontrol,MD} to the panoramic domain and establish a new benchmark. We conduct a comprehensive comparison between Percep360 and other state-of-the-art approaches. In the NRIQA evaluation, Percep360 achieves the best performance, particularly in the PIQE metric, where our method surpasses the original stitched images. Visualization results, as shown in Fig.~\ref{fig:Fig4}, further demonstrate that the stitched images suffer from obvious pixel inconsistencies, whereas our generated images effectively resolve this issue and produce more coherent results. Other methods also exhibit a tendency to learn the misalignment present in the ground truth during training, which leads to incoherent image generation in their visualization outputs. Regarding IQA and controllability, our network also attains strong performance. For MagicDrive, we modify its cross-view attention to self-attention, better suited for panoramic images (an advantage of panoramic imaging that eliminates the need for explicit viewpoint alignment). Although Percep360’s FID is slightly higher than that of MagicDrive, it achieves the best NRIQA, controllability, and overall score in the comprehensive evaluation.

\section{Ablation Studies}

The effectiveness of our method is verified through ablation studies on its two main components. Coherent panorama generation strategies (LSDM) are compared in Sect.~\ref{Sec.VA}, followed by an evaluation of the performance of prompt utilization in Sec~\ref{Sec.VB}. We further validate that the generated data contributes to the improvement of downstream BEV segmentation performance in Sec.~\ref{Sec.VC}.

\begin{table*}[htbp]
    \centering
    \begin{minipage}[t]{0.93\textwidth}
        \vspace{0pt} 
        \centering
        \resizebox{\textwidth}{!}{
            \setlength{\tabcolsep}{8pt}
            \begin{tabular}{l|cc|c|cc|c}
            \toprule
            Settings & BRISQUE~\cite{BRISQUE}$\downarrow$ & PIQE~\cite{PIQE}$\downarrow$ & FID~\cite{FID}$\downarrow$ & Driv.$\uparrow$ & Mean.$\uparrow$ & Rank.$\downarrow$ \\
            \midrule
            
            Baseline & 22.41 (5) & 14.33 (5) & 22.07 (5) & 0.21 (1) & 0.12 (1) & 17 (3) \\
            + Mask & 20.87 (4) & 11.65 (4) & 16.35 (3) & 0.18 (3) & 0.09 (3) & 17 (4) \\
            + M.Crs & 20.03 (1) & 10.91 (2) & 14.96 (1) & 0.17 (4) & 0.08 (4) & 12 (1) \\
            + Rota & 20.25 (2) & 10.85 (1) & 21.42 (4) & 0.14 (5) & 0.06 (5) & 17 (5) \\
            + LSDM F & 20.84 (3) & 11.44 (3) & 16.28 (2) & 0.20 (2) & 0.10 (2) & 12 (1) \\
            \midrule
            \bottomrule
        \end{tabular}

        }
        \vspace{1.5pt}

    \end{minipage}
    \hfill
    \caption{An in-depth analysis of LSDM. $\uparrow$/$\downarrow$ indicates whether a higher or lower value is preferred for each metric. ``Mask'' refers to the masking strategy, while ``M.Crs.'' denotes the cross-attention method built upon the masking strategy. Additionally, ``Rota'' represents the rotation method. ``Driv.'' refers to the segmentation results of drivable area, and ``Mean'' denotes the average performance across all BEV segmentation classes.} 
    \label{tab:LSDM}
    \vspace{-1em}
\end{table*}

\begin{table*}[htbp]
    \centering
    \begin{minipage}[t]{0.93\textwidth}
        \vspace{0pt} 
        \centering
        \resizebox{\textwidth}{!}{
            \setlength{\tabcolsep}{8pt}
            \begin{tabular}{l|cc|c|cc|c}
            \toprule
                Settings    & BRISQUE~\cite{BRISQUE}$\downarrow$ & PIQE~\cite{PIQE}$\downarrow$ & FID~\cite{FID}$\downarrow$ & Driv.$\uparrow$ & Mean.$\uparrow$ & Rank.$\downarrow$ \\
                \midrule
                
                Baseline       & 22.41 (5) & 14.33 (5) & 22.07 (5) & 0.21 (4) & 0.12 (2) & 21 (5) \\
                LSDM          & 20.87 (3) & 12.25 (4) & 16.28 (4) & 0.20 (5) & 0.10 (5) & 21 (5) \\
                LSMD + D      & 20.41 (2) & 11.54 (2) & 12.76 (1) & 0.23 (2) & 0.12 (2) & 9  (2) \\
                Percep360    & 20.24 (1) & 11.44 (1) & 14.43 (3) & 0.25 (1) & 0.13 (1) & 7  (1) \\ 
                \bottomrule \midrule
            \end{tabular}
        }
        \vspace{1.5pt}
    \end{minipage}
    \hfill
    \caption{Ablation results under different prompt configurations. ``D'' refers to the Depth map. ``Driv.'' refers to the segmentation results of drivable area, and ``Mean'' denotes the average performance across all BEV segmentation classes.}
    \label{tab:ablation}
    \vspace{-2em}
\end{table*}

\subsection{Analysis of LSDM}\label{Sec.VA} To address the inherent incoherence in scene capture arising from multi-view camera systems during data collection, we compared LSDM against three conventional approaches, as shown in Tab.~\ref{tab:LSDM}. It is worth noting that, to the best of our knowledge, there has been no prior discussion in the fields of panoramic and street-view image generation regarding the relationship between 3D projection into the BEV space and image coherence within the BEV view. First,
the masking strategy (``Mask'')~\cite{StoryGen,LFC,RePaint} serves as a training guidance mechanism within the diffusion paradigm. By intentionally omitting supervision in suboptimal or uncertain regions, the model is encouraged to leverage contextual information from surrounding areas to reconstruct or complete the masked content, thereby enhancing its reconstruction capabilities. Furthermore, we incorporate self-attention (``M.Crs'') over latent features within the U-Net architecture to enhance the diffusion of supervision signals from observed to unobserved regions, resulting in noticeable improvements in image quality at the feature level. However, those methods are not suitable for the present task. As street view generation requires road and geometric details, methods that rely solely on neighboring features, although effective in preserving semantic consistency, lead to a significant deterioration in the structural integrity of the masked regions. Furthermore, following the training paradigm of state-of-the-art panoramic generation methods~\cite{Panodiff1,360dvd}, we rotate the image pixels and apply the corresponding top-down rotation to the BEV image. However, this approach results in a significant degradation in both image quality and layout controllability. LSDM decouples the correspondence between the image-BEV pair and dynamically reconfigures the relationship between the network and these pairs. This approach effectively mitigates the erroneous constraints introduced by sample incoherence from pinhole cameras, while leveraging the model’s learning capacity to alleviate the controllability degradation caused by the other methods. We further conduct a ranking of various image evaluation metrics, demonstrating that LSDM effectively achieves a balance between image quality and layout controllability.

\begin{table}[hbp]
    \centering

    \resizebox{0.48\textwidth}{!}{
    \begin{tabular}{l|cc}
        \toprule
        \multirow{2}{*}{Data} & \multicolumn{2}{c}{BEV Segmentation} \\ \cline{2-3}
                              & Driv. ($\uparrow$) & Mean. ($\uparrow$) \\
        \midrule
        w/o synthetic data &0.593 &0.470 \\
        w/ Baseline &0.589 &0.469 \\
        \rowcolor{dcolor} w/ Percep360 & 0.610&0.495 \\
        \bottomrule
    \end{tabular}
    }
    \vspace{0.5em}
    \caption{
       By leveraging BEV features along with additional conditioning signals as inputs, new panoramic street-view images are synthesized through different methods. The generated data is further used to boost the panoramic BEV segmentation task.
    }
    \label{tab:support_train}
    \vspace{-2em}
\end{table}

\subsection{Analysis of Different Prompts}\label{Sec.VB} To achieve more controllable panoramic image generation and avoid the labor-intensive task of manual panoramic image annotation, we leverage large language models to pre-annotate the stitched panoramic images, thereby generating imprecise yet informative control prompts. Furthermore, we propose the PPM, which enables robust panoramic generation. By incorporating depth information, PPM dynamically selects relevant image tokens and depth cues, surpassing the baseline in terms of image quality while significantly improving controllability.  With the joint contributions of PPM and LSMD, Percep360 achieves comprehensive improvements in both image quality and layout controllability.

\subsection{Training Support for BEV Segmentation.}\label{Sec.VC} As an important validation of the capabilities of the world model, we evaluate the effectiveness of newly generated data as augmentation for improving downstream network performance. Specifically, we adopt OneBEV~\cite{Onebev} as the baseline for BEV segmentation. As shown in Table \ref{tab:support_train}, incorporating synthetic data into the baseline model did not lead to performance improvement (mIoU dropped slightly from $0.470$ to $0.469$). In contrast, the synthetic images generated by Percep360 exhibit higher realism and better alignment with BEV representations, resulting in performance gains: the drivable region improved from $0.593$ to $0.610$, and the overall mIoU increased from $0.470$ to $0.495$. Notably, these benefits were achieved without the need for additional data cleaning or filtering, making Percep360 a cost-effective paradigm for enhancing agent perception.

\section{Conclusion and Future Work}
\noindent\textit{Contribution:} In this paper, we propose the first panoramic generation method Percep360. By introducing the Local Scenes Diffusion Method, the panorama generation task is reformulated as a spatially continuous diffusion process, ensuring the coherence of panoramic images. Additionally, the Probabilistic Prompting Method dynamically selects the most relevant control cues, thereby improving the controllability of image generation. The generated data proves to be an effective form of data augmentation, leading to improved performance in downstream tasks.

\noindent\textit{Limitation and Future Work:} In this paper, we focus on exploiting image coherence and the controllability of 2D features. The label mapping from six-view images to panoramic representations constitutes a critical component for efficient panoramic data generation. Furthermore, existing street-view generation approaches have largely overlooked the effective utilization of prompt features, which we identify as another key direction for future investigation. 
%

\bibliographystyle{IEEEtran}
\bibliography{1}

@article{AutoPano,
  title={Review on panoramic imaging and its applications in scene understanding},
  author={Gao, Shaohua and Yang, Kailun and Shi, Hao and Wang, Kaiwei and Bai, Jian},
  journal={IEEE Transactions on Instrumentation and Measurement},
  volume       = {71},
  pages        = {1--34},
  year={2022},
  publisher={IEEE}
}

@article{PanoramicReview2,
  title={Spherical {DNNs} and Their Applications in 360{\textdegree} Images and Videos},
  author={Xu, Yanyu and Zhang, Ziheng and Gao, Shenghua},
  journal={IEEE Transactions on Pattern Analysis and Machine Intelligence},
  volume       = {44},
  number       = {10},
  pages        = {7235--7252},
  year         = {2022},
  publisher={IEEE}
}

@article{ai2025survey,
  title={A survey of representation learning, optimization strategies, and applications for omnidirectional vision},
  author={Ai, Hao and Cao, Zidong and Wang, Lin},
  journal={International Journal of Computer Vision},
  volume       = {133},
  number       = {8},
  pages        = {4973--5012},
  year={2025}
}

@article{Panoflow,
  title={{PanoFlow:} {Learning} 360{\textdegree} Optical Flow for Surrounding Temporal Understanding},
  author={Shi, Hao and others},
  journal={IEEE Transactions on Intelligent Transportation Systems},
  volume       = {24},
  number       = {5},
  pages        = {5570--5585},
  year={2023},
  publisher={IEEE}
}

@inproceedings{Panoflow2,
  title={Deep 360{\textdegree} optical flow estimation based on multi-projection fusion},
  author={Li, Yiheng and Barnes, Connelly and Huang, Kun and Zhang, Fang-Lue},
  booktitle={ECCV},
  pages={336--352},
  year={2022}
}

@inproceedings{PanoOD,
  title={{PANDORA:} {A} Panoramic Detection Dataset for Object with Orientation},
  author={Xu, Hang and others},
  booktitle={ECCV},
  pages={237--252},
  year={2022}
}

@inproceedings{Ominitrack,
  title={Omnidirectional Multi-Object Tracking},
  author={Luo, Kai and others},
  booktitle={CVPR},
  pages={21959--21969},
  year={2025}
}

@inproceedings{PanoSS3,
  title={Semantics distortion and style matter: Towards source-free {UDA} for panoramic segmentation},
  author={Zheng, Xu and Zhou, Pengyuan and Vasilakos, Athanasios V and Wang, Lin},
  booktitle={CVPR},
  pages        = {27885--27895},
  year={2024}
}

@inproceedings{PanoSS,
  title={Open panoramic segmentation},
  author={Zheng, Junwei and others},
  booktitle={ECCV},
  pages={164--182},
  year={2024}
}

@inproceedings{Onebev,
  title={{OneBEV:} {Using} One Panoramic Image for Bird's-Eye-View Semantic Mapping},
  author={Wei, Jiale and Zheng, Junwei and Liu, Ruiping and Hu, Jie and Zhang, Jiaming and Stiefelhagen, Rainer},
  booktitle={ACCV},
  pages={583--596},
  year={2024}
}

@inproceedings{ControlNet,
  title={Adding conditional control to text-to-image diffusion models},
  author={Zhang, Lvmin and Rao, Anyi and Agrawala, Maneesh},
  booktitle={ICCV},
  pages        = {3813--3824},
  year={2023}
}

@article{Controlnext,
  title={{ControlNeXt:} {Powerful} and Efficient Control for Image and Video Generation},
  author={Peng, Bohao and Wang, Jian and Zhang, Yuechen and Li, Wenbo and Yang, Ming-Chang and Jia, Jiaya},
  journal={arXiv preprint arXiv:2408.06070},
  year={2024}
}

@article{BevGen,
  title={Street-view image generation from a bird's-eye view layout},
  author={Swerdlow, Alexander and Xu, Runsheng and Zhou, Bolei},
  journal={IEEE Robotics and Automation Letters},
  year={2024},
  volume={9},
  number={4},
  pages={3578--3585},
  publisher={IEEE}
}

@article{Bevcontrol,
  title={{BEVControl:} {Accurately} Controlling Street-view Elements with Multi-perspective Consistency via {BEV} Sketch Layout},
  author={Yang, Kairui and Ma, Enhui and Peng, Jibin and Guo, Qing and Lin, Di and Yu, Kaicheng},
  journal={arXiv preprint arXiv:2308.01661},
  year={2023}
}

@inproceedings{Drive123,
  title={Driving into the future: Multiview visual forecasting and planning with world model for autonomous driving},
  author={Wang, Yuqi and He, Jiawei and Fan, Lue and Li, Hongxin and Chen, Yuntao and Zhang, Zhaoxiang},
  booktitle={CVPR},
  pages={14749--14759},
  year={2024}
}

@inproceedings{Occword,
  title={{OccWorld:} {Learning} a {3D} Occupancy World Model for Autonomous Driving},
  author={Zheng, Wenzhao and Chen, Weiliang and Huang, Yuanhui and Zhang, Borui and Duan, Yueqi and Lu, Jiwen},
  booktitle={ECCV},
  pages={55--72},
  year={2025},
}

@inproceedings{EdiDriving,
  title={Editable Scene Simulation for Autonomous Driving via Collaborative {LLM}-Agents},
  author={Wei, Yuxi and others},
  booktitle={CVPR},
  pages        = {15077--15087},
  year={2024}
}

@inproceedings{MD,
  title={{MagicDrive:} {Street} View Generation with Diverse {3D} Geometry Control},
  author={Gao, Ruiyuan and others},
  booktitle={ICLR},
  year={2024}
}

@inproceedings{LDM,
  title={High-resolution image synthesis with latent diffusion models},
  author={Rombach, Robin and Blattmann, Andreas and Lorenz, Dominik and Esser, Patrick and Ommer, Bj{\"o}rn},
  booktitle={CVPR},
  pages        = {10674--10685},
  year={2022}
}

@article{SSIM,
  title={Image quality assessment: from error visibility to structural similarity},
  author={Wang, Zhou and Bovik, Alan C and Sheikh, Hamid R and Simoncelli, Eero P},
  journal={IEEE Transactions on Image Processing},
  volume={13},
  number={4},
  pages={600--612},
  year={2004},
  publisher={IEEE}
}

@article{BRISQUE,
  title={No-reference image quality assessment in the spatial domain},
  author={Mittal, Anish and Moorthy, Anush Krishna and Bovik, Alan Conrad},
  journal={IEEE Transactions on Image Processing},
  volume={21},
  number={12},
  pages={4695--4708},
  year={2012},
  publisher={IEEE}
}

@inproceedings{FID,
  title={{GANs} Trained by a Two Time-Scale Update Rule Converge to a Local Nash Equilibrium},
  author={Heusel, Martin and Ramsauer, Hubert and Unterthiner, Thomas and Nessler, Bernhard and Hochreiter, Sepp},
  booktitle={NeurIPS},
  pages        = {6626--6637},
  year={2017}
}

@inproceedings{PIQE,
  title={Blind image quality evaluation using perception based features},
  author={Venkatanath N. and
                  Praneeth D. and
                  Maruthi Chandrasekhar Bh. and
                  Sumohana S. Channappayya and
                  Swarup S. Medasani},
  booktitle={NCC},
  pages={1--6},
  year={2015}
}

@inproceedings{Nuscenes,
  title={{nuScenes:} {A} Multimodal Dataset for Autonomous Driving},
  author={Caesar, Holger and others},
  booktitle={CVPR},
  pages        = {11618--11628},
  year={2020}
}

@inproceedings{StoryGen,
  title={Intelligent Grimm - {Open-ended} Visual Storytelling via Latent Diffusion Models},
  author={Liu, Chang and Wu, Haoning and Zhong, Yujie and Zhang, Xiaoyun and Wang, Yanfeng and Xie, Weidi},
  booktitle={CVPR},
  pages={6190--6200},
  year={2024}
}

@inproceedings{LFC,
  title={Resolution-robust large mask inpainting with fourier convolutions},
  author={Suvorov, Roman and others},
  booktitle={WACV},
  pages        = {3172--3182},
  year={2022}
}

@inproceedings{RePaint,
  title={{RePaint:} {Inpainting} using Denoising Diffusion Probabilistic Models},
  author={Lugmayr, Andreas and Danelljan, Martin and Romero, Andres and Yu, Fisher and Timofte, Radu and Van Gool, L Repaint},
  booktitle={CVPR},
   pages={11451--11461},
  year={2022}
}

@inproceedings{Panodiff1,
  title={360-Degree Panorama Generation from Few Unregistered {NFoV} Images},
  author={Wang, Jionghao and Chen, Ziyu and Ling, Jun and Xie, Rong and Song, Li},
  booktitle={MM},
  pages={6811--6821},
  year={2023}
}

@inproceedings{360dvd,
  title={{360DVD:} {Controllable} panorama video generation with 360-degree video diffusion model},
  author={Wang, Qian and Li, Weiqi and Mou, Chong and Cheng, Xinhua and Zhang, Jian},
  booktitle={CVPR},
  pages={6913--6923},
  year={2024}
}

@article{DriveScape,
  title={{DriveScape:} {Towards} High-Resolution Controllable Multi-View Driving Video Generation},
  author={Wu, Wei and others},
  journal={arXiv preprint arXiv:2409.05463},
  year={2024}
}

@inproceedings{DrivingDiffusion,
  title={{DrivingDiffusion:} {Layout-guided} Multi-view Driving Scenarios Video Generation with Latent Diffusion Model},
  author={Li, Xiaofan and Zhang, Yifu and Ye, Xiaoqing},
  booktitle={ECCV},
  pages={469--485},
  year={2024}
}

@article{holodrive,
  title={{HoloDrive:} {Holistic} {2D-3D} {Multi-modal} Street Scene Generation for Autonomous Driving},
  author={Wu, Zehuan and others},
  journal={arXiv preprint arXiv:2412.01407},
  year={2024}
}

@inproceedings{Tex2St,
  title={{Text2Street:} {Controllable} Text-to-Image Generation for Street Views},
  author={Gu, Songen and Su, Jinming and Duan, Yiting and Chen, Xingyue and Luo, Junfeng and Zhao, Hao},
  booktitle={ICPR},
  pages={130--145},
  year={2025}
}

@inproceedings{panacea,
  title={{Panacea:} {Panoramic} and controllable video generation for autonomous driving},
  author={Wen, Yuqing and others},
  booktitle={CVPR},
  pages={6902--6912},
  year={2024}
}

@article{Magicdrivedit,
  title={{MagicDriveDiT:} {High-resolution} Long Video Generation for Autonomous Driving with Adaptive Control},
  author={Gao, Ruiyuan and Chen, Kai and Xiao, Bo and Hong, Lanqing and Li, Zhenguo and Xu, Qiang},
  journal={arXiv preprint arXiv:2411.13807},
  year={2024}
}

@article{Perldiff,
  title={{PerlDiff:} {Controllable} Street View Synthesis Using Perspective-Layout Diffusion Models},
  author={Zhang, Jinhua and others},
  journal={arXiv preprint arXiv:2407.06109},
  year={2024}
}

@article{UniMLVG,
  title={{UniMLVG:} {Unified} Framework for Multi-view Long Video Generation with Comprehensive Control Capabilities for Autonomous Driving},
  author={Chen, Rui and others},
  journal={arXiv preprint arXiv:2412.04842},
  year={2024}
}

@article{Stag1,
  title={{Stag-1:} {Towards} Realistic {4D} Driving Simulation with Video Generation Model},
  author={Wang, Lening and others},
  journal={arXiv preprint arXiv:2412.05280},
  year={2024}
}

@inproceedings{Dreammatcher,
  title={{DreamMatcher:} {Appearance} Matching Self-Attention for Semantically-Consistent Text-to-Image Personalization},
  author={Nam, Jisu and Kim, Heesu and Lee, DongJae and Jin, Siyoon and Kim, Seungryong and Chang, Seunggyu},
  booktitle={CVPR},
   pages={8100--8110},
  year={2024}
}

@inproceedings{OPT2I,
  title={Optimizing prompts for text-to-image generation},
  author={Hao, Yaru and Chi, Zewen and Dong, Li and Wei, Furu},
  booktitle={NeurIPS},
  pages={66923--66939},
  year={2023}
}

@inproceedings{Raphael,
  title={Raphael: Text-to-image generation via large mixture of diffusion paths},
  author={Xue, Zeyue and others},
  booktitle={NeurIPS},
  pages={41693--41706},
  year={2024}
}

@inproceedings{PanFusion,
  title={Taming Stable Diffusion for Text to 360{\textdegree} Panorama Image Generation},
  author={Zhang, Cheng and others},
  booktitle={CVPR},
  pages={6347--6357},
  year={2024}
}

@inproceedings{PanoFree,
  title={{PanoFree:} {Tuning-free} Holistic Multi-view Image Generation with Cross-View Self-guidance},
  author={Liu, Aoming and Li, Zhong and Chen, Zhang and Li, Nannan and Xu, Yi and Plummer, Bryan A},
  booktitle={ECCV},
  pages={146--164},
  year={2024}
}

@inproceedings{PanoGen,
  title={{PanoGen:} {Text-conditioned} panoramic environment generation for vision-and-language navigation},
  author={Li, Jialu and Bansal, Mohit},
  booktitle={NeurIPS},
  pages={21878--21894},
  year={2023}
}

@article{PICFormer,
  title={Bridging Global Context Interactions for High-Fidelity Pluralistic Image Completion},
  author={Zheng, Chuanxia and Song, Guoxian and Cham, Tat-Jen and Cai, Jianfei and Luo, Linjie and Phung, Dinh},
  journal={IEEE Transactions on Pattern Analysis and Machine Intelligence},
  year={2024},
  volume={46},
  number={12},
  pages={8320--8333},
  publisher={IEEE}
}

@inproceedings{PIC,
  title={Pluralistic image completion},
  author={Zheng, Chuanxia and Cham, Tat-Jen and Cai, Jianfei},
  booktitle={CVPR},
  pages={1438--1447},
  year={2019}
}

@inproceedings{Brushnet,
  title={{BrushNet:} {A} Plug-and-Play Image Inpainting Model with Decomposed Dual-Branch Diffusion},
  author={Ju, Xuan and Liu, Xian and Wang, Xintao and Bian, Yuxuan and Shan, Ying and Xu, Qiang},
  booktitle={ECCV},
  pages={150--168},
  year={2024}
}

@article{Brushedit,
  title={{BrushEdit:} {All-in-one} Image Inpainting and Editing}, 
  author={Li, Yaowei and Bian, Yuxuan and Ju, Xuan and Zhang, Zhaoyang and Shan, Ying and Xu, Qiang},
  journal={arXiv preprint arXiv:2412.10316},
  year={2024}
}

@inproceedings{PanoDiff2,
  title={{PanoDiffusion:} {360-degree} Panorama Outpainting via Diffusion},
  author={Wu, Tianhao and Zheng, Chuanxia and Cham, Tat-Jen},
  booktitle={ICLR},
  year={2023}
}

@inproceedings{PALP,
  title={{PALP:} {Prompt} Aligned Personalization of Text-to-Image Models},
  author={Arar, Moab and others},
  booktitle={SIGGRAPH Asia},
  pages={1--11},
  year={2024}
}

@inproceedings{clip,
  title={Learning transferable visual models from natural language supervision},
  author={Radford, Alec and others},
  booktitle={ICML},
  pages={8748--8763},
  year={2021},
}

@article{DriveSim,
  title={Learning a driving simulator},
  author={Santana, Eder and Hotz, George},
  journal={arXiv preprint arXiv:1608.01230},
  year={2016}
}

\end{document}